\documentclass{article}

     \usepackage[final]{neurips_2020}

\usepackage[utf8]{inputenc} 
\usepackage[T1]{fontenc}    
\usepackage{hyperref}       
\usepackage{url}            
\usepackage{booktabs}       
\usepackage{amsfonts}       
\usepackage{nicefrac}       
\usepackage{microtype}      
\usepackage{graphicx}       

\title{Linear programming word problems formulation using EnsembleCRF NER labeler and T5 text generator with data augmentations}

\author{%
  JiangLong He, Mamatha N, Shiv Vignesh, Deepak Kumar, Akshay Uppal\\
  Infrrd AI Lab\\
  Infrrd.ai\\
  \texttt{\{jianglong,mamathan,shivvignesh,deepakumar,akshayuppal\}@infrrd.ai} \\
}

\begin{document}

\maketitle

\begin{abstract}
We propose an ensemble approach to predict the labels in linear programming word problems. The entity identification and the meaning representation are two types of tasks to be solved in the NL4Opt competition. We propose the ensembleCRF method to identify the named entities for the first task. We found that single models didn't improve for the given task in our analysis. A set of prediction models predict the entities. The generated results are combined to form a consensus result in the ensembleCRF method. We present an ensemble text generator to produce the representation sentences for the second task. We thought of dividing the problem into multiple small tasks due to the overflow in the output. A single model generates different representations based on the prompt. All the generated text is combined to form an ensemble and produce a mathematical meaning of a linear programming problem.
\end{abstract}

\section{Introduction}

Linear programming problems are used in business applications to optimize the solution found for the problem. We define the linear programming problem as a maximization or minimization problem. We may have to obtain maximum profit or reduce operational costs based on the requirements. People conceived the idea of formulating a mathematical problem to run a business and drive society to the next level. The problem arises when there are multiple ingredients or resources which scale only in a linear manner. We like to blend two or more materials to maximize profit or minimize cost, an example of linearity in the materialistic view.

The formulation of linear programming problems is a powerful concept in solving a set of business problems consisting of monetary benefits. The solution for the linear programming problems is still a critical concept to understand by non-experts. Dantzig developed a method known as a simplex method to solve linear programming problems \cite{nash}. Here a set of constraints defines the feasible region to optimize the linear programming problem. We locate the corners of the feasible region and evaluate them in this method to find the optimal solution. The solutions to the linear programming problems got better with interior point methods. In an interior point-based approach, the valuation begins from inside the feasible region, which reduces the computational time compared to the simplex method when the number of variables is large \cite{karmarkar}. Now, we have readily available software to solve the problem \cite{cplex}. The software helps to solve mathematically framed linear programming problems. It still requires an expert to formulate a linear programming problem from the word problem. The NL4Opt competition \cite{nl4optcomp} creates an opportunity to develop an automated solution to convert a word problem into a mathematically framed problem which is then passed on to readily available software to find the optimal solution \cite{nl4opt}. The competition conceptualizes the hurdle faced by the non-experts while formulating a linear programming problem. There are two tasks in this competition, and they are named entity recognition and generating the precise meaning representation. A method has to identify all the entities present in the word problems in the entity recognition task. Another solution uses the located entities to generate a mathematical representation from the previous method. However, the text generation method depends on the ground truth of the first task and acts as a reference for evaluation.

Our contributions and observations are summarized as follows:
\begin{itemize}
\item The number of samples in the training set is less to train a large language model. So, we introduce the data augmentation strategies: LwTR, SR, MR, and SiS \cite{dai2020analysis} as a preprocessing step. Thus, we increase the number of samples in the data to provide a more plausible output.
\item We introduce an ensemble of models to learn the weights for each label predicted from the single models. 
\item We introduce multi-task learning and train a large language model to generate text through different prompts. An entity wrapper is placed around an entity to enhance the data. We combine the task-specific text from the same model to form a meaningful representation for linear programming solvers.
\end{itemize}

\section{Sub-task 1: named entity recognition}

\begin{figure}[b]
  \centering
  \includegraphics[scale=0.25]{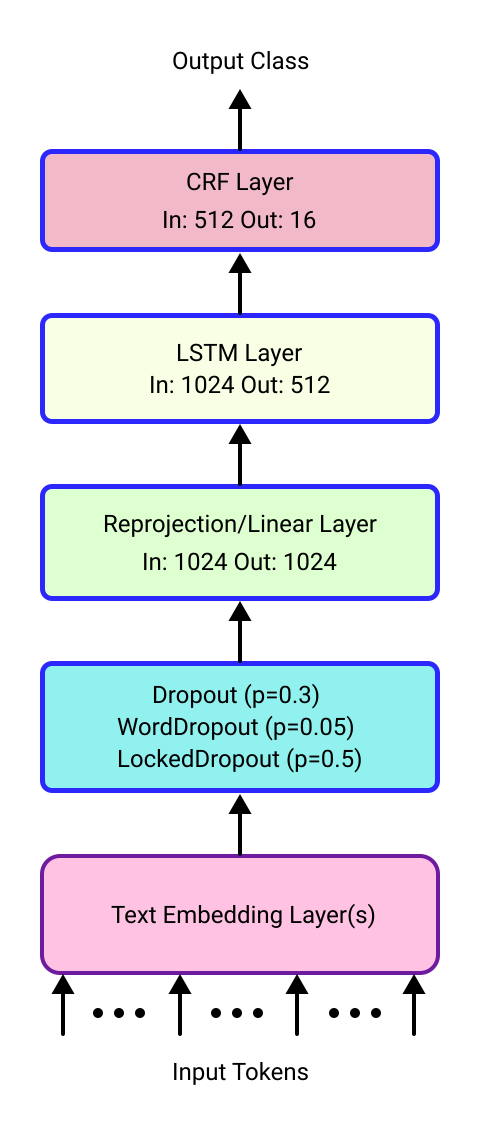}
\caption{The main architecture of the model where the text embedding layer is replaced with different types of embeddings available in hugging face.}
\label{fig1}
\end{figure}

Our approach follows the flair framework \cite{akbik2019flair} and uses PyTorch and hugging face \cite{huggingface} transformers to build and experiment. Our base model is a ‘Text Embedding + BiLSTM + CRF’ transformer model as shown in Figure \ref{fig1}. We had used the base model to train a multilingual transformer for MultiCONER competition \cite{multiconer-data, multiconer-report}. The text embedding layer was XLM-Roberta embeddings for MultiCONER competition \cite{he-etal-2022-infrrd}. The F1 score of the model is equivalent to the baseline results when the text embedding layer is Roberta-base \cite{liu2019roberta}. Since the number of training samples available to train the model is less, we performed data augmentation techniques on the training samples.

\subsection{Data augmentation}

A lot of labeled data is required by transformer-based language models to produce a good performance. Here, we are working on linear programming word problems, and the availability of the word problems is scarce. Apart from this, we have a time-consuming annotation task that requires expert knowledge. In this case, both of them are very critical. We need to overcome this data deficiency problem. So, we made use of data augmentation techniques to replace token-level embeddings \cite{dai2020analysis}. The method expands the number of training samples by adding additional transformed samples without changing the labels. There are four techniques in this method: Label-wise token replacement (LwTR), Synonym replacement (SR), Mention replacement (MR), and Shuffle within segments (SiS). Here, we have used all the techniques to perform data augmentation in the training samples. However, the development set was not utilized and kept as it was for evaluation purposes. Table \ref{tab1} illustrates an example of generation of augmented data.

\begin{table*}[h]
    \begin{center}
    \caption{The different types of data augmentation techniques: Label-wise token replacement (LwTR), Synonym replacement (SR), Mention replacement (MR), and Shuffle within segments (SiS).}
    \label{tab1}
    \begin{tabular}{l|llllllll} 
      \hline
      Method & Instance & & & & & & \\
      \hline
      \hline
      Original & A & serving & of & chicken &  & costs & \$ & 10\\
      & O & O & O & B-VAR & & B-OBJ\_NAME & O & B-PARAM\\
      \hline
      LwTR & A & flour & of & cereal & & profit & \$ & 25\\
      & O & O & O & B-VAR & & B-OBJ\_NAME & O & B-PARAM\\
      \hline
      SR & A & serving & of & volaille & & cost & \$ & X\\
      & O & O & O & B-VAR & & B-OBJ\_NAME & O & B-PARAM\\
      \hline
      MR & A & serving & of & Durian & TV & savings & \$ & 0.32\\
      & O & O & O & B-VAR & I-VAR & B-OBJ\_NAME & O & B-PARAM\\
      \hline
      SiS & serving & of & a & chicken & & costs & \$ & 25\\
      & O & O & O & B-VAR & & B-OBJ\_NAME & O & B-PARAM\\
      \hline
    \end{tabular}
    \end{center}
\end{table*}

\subsubsection{Single model accuracy after data augmentation}

We started evaluating the models with different types of embeddings and a combination of embeddings. The results are tabulated in Table \ref{tab2}. The XLM-Roberta-base model \cite{conneau2019unsupervised} was the initial choice as experimented with in the baseline results. Quickly, we moved to the next set of embeddings due to an increase in the F1 score for different types of embeddings. The Roberta-base embeddings provided better results compared to the XLM-Roberta-base. We performed ablation study by changing the text embeddings and their combination. The transformer model has complex contextual information between the tokens. Whether the contextual information from the transformer model is sufficient or not is evaluated in the single model by removing a few of the components in the pipeline. In the seventh and eighth rows, nobilstm and nobilstmcrf refer to without the BiLSTM module and BiLSTM+CRF module from the model. The F1 score was better without the BiLSTM module giving a glimpse of contextual information gained from the transformer rather than the separate BiLSTM module. However, we couldn't distinguish a single model that is better than other. 

\begin{table*}[h]
    \begin{center}
    \caption{The performance of a single model with different text embeddings on the development set.}
    \label{tab2}
    \begin{tabular}{l|l|l} 
      \hline
      Model& Text embeddings & Development\\
      number & & F1 score\\
      \hline
      \hline
      1 & roberta-base + no augmentation & 0.8117\\
      \hline
      2 & roberta-base + LwTR \& SR augmentation & 0.8641\\
      \hline
      3 & xlm-roberta-base + 4 augmentations & 0.8898\\
      \hline
      4 & roberta-base + 4 augmentations & 0.9127\\
      \hline
      5 & roberta-base + glove + 4 augmentations & 0.9151\\
      \hline
      6 & only glove + 4 augmentations & 0.787\\
      \hline
      7 & roberta-base + glove + nobilstm + 4 augmentations & 0.9154\\
      \hline
      8 & roberta-base + glove + nobilstmcrf + 4 augmentations & 0.9034\\
      \hline
    \end{tabular}
    \end{center}
\end{table*}

\subsection{Ensemble architecture}

A simple ensemble strategy of \textit{Majority Voting} is developed by \cite{he-etal-2022-infrrd}. Given a set of M sequence labeling models denoted as $C = \{ c_1, c_2, ..., c_M \}$ and an input sentence denoted as $S = \{ w_1, w_2, ..., w_n \}$, where each $w$ is a word from S. Each model from $C$ will output a sequence of predictions for each word $w$ in sentence $S$. Let $O^{s}_{c_i} = \{ O^{c_i}_{w_1}, O^{c_i}_{w_2}, ..., O^{c_i}_{w_n} \}$ denote the prediction output of model $C_i$ on sentence $S$. $O^{c_i}_{w_j}$ denotes the prediction of model $C_i$ on word $w_j$ in IOB format. The set of outputs for all models in $C$  on sentence $S$ will be denoted as

\begin{equation}
O_S = \{ O^{s}_{c_1}, O^{s}_{c_2}, ..., O^{s}_{c_M}  \}
\end{equation}

\begin{figure}[h]
\centering
\includegraphics[scale=0.16]{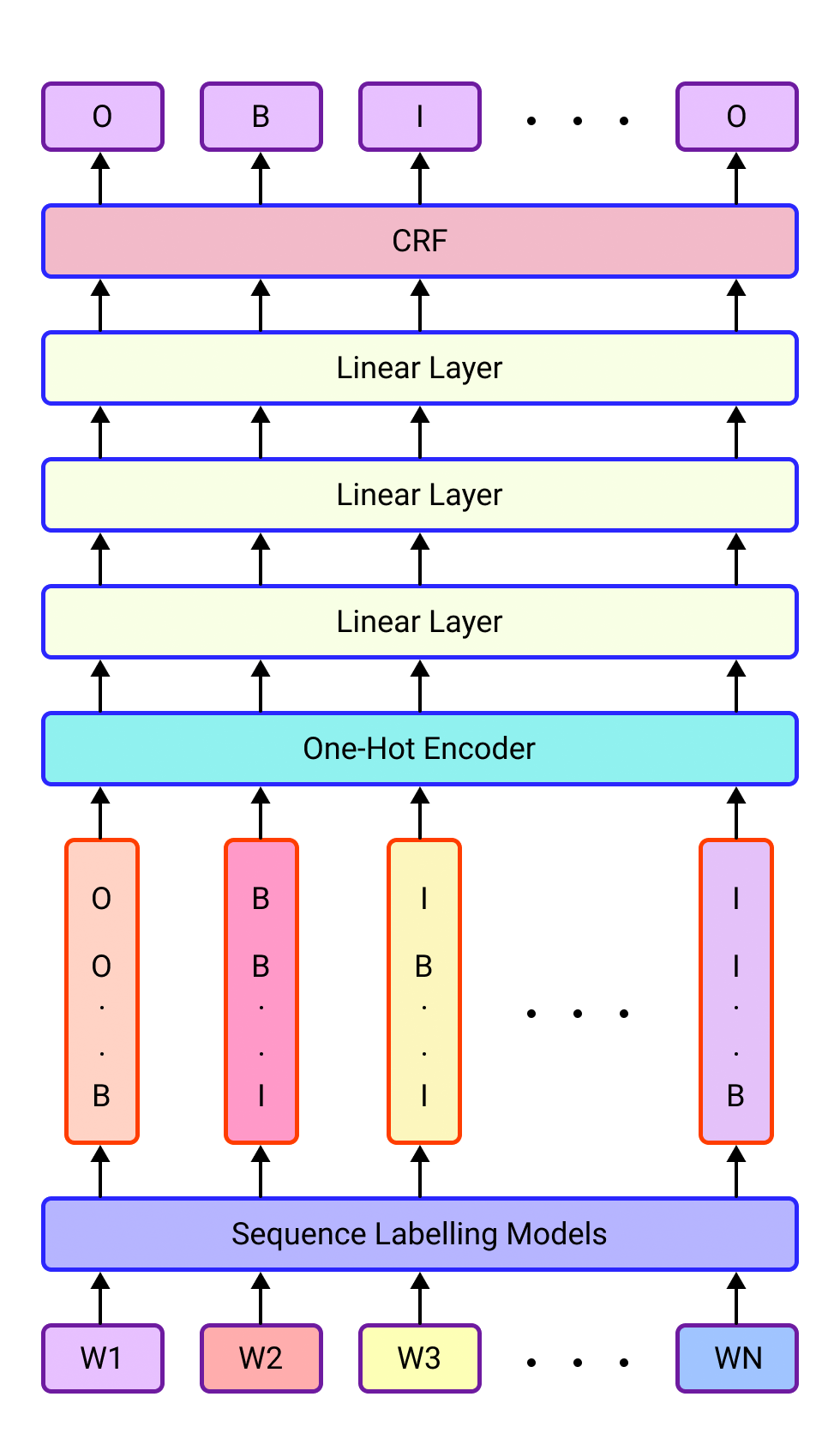}
\caption{The architecture of EnsembleCRF model. Given a sentence as input, each of the sequence labeling models will output the name entity prediction in IOB format. One hot encoder combines them to generate ensemble output from a Conditional Random Field model.}
\label{fig2}
\end{figure}

The \textit{Majority Voting} strategy takes all model’s predictions of word $w_j$ and outputs the most frequent prediction as to the final prediction for $w_j$. An obvious issue with \textit{Majority Voting} is the IOB scheme constraint. The final ensemble result is not guaranteed to be passing all the constraints where (I) tag must follow and (B) tag and the entity of neighbor (B) and (I) tag must be the same. We introduce an ensemble learning approach via sequence labeling called `EnsembleCRF' as shown in Figure \ref{fig2}.

The model outputs are stacked together and passed through a one-hot encoder, three linear layers, and CRF. The CRF layer is trained to optimally combine the model predictions to form a new set of predictions. The addition of the three linear layers helped in the performance improvement. The EnsembleCRF model is of the form

\begin{equation}
\begin{array}{c}
C_{en} = EnsembleCRF(C = \{ c_1, c_2, ..., c_M \},\\
D_{en}=\{ X,Y\})
\end{array}
\end{equation}

$D_{en}$ is the ensemble learning dataset composed of $X$ and $Y$. $X = \{ O_{s_1}, O_{s_2}, ..., O_{s_k} \}$ is created by using model set $C$ and set of input sentences $\{S\} = \{ S_1, S_2, ..., S_k\}$ with size K. Each element of $X$ is defined as equation (1). $Y$ is the ground truth entity label in IOB format. During the training phase for some models in set $C$, we included only the training datasets. Thus, we choose to perform the ensemble learning using an augmented train set created using the data augmentation strategies explained in Section 2.1. Since the second layer classifier is a CRF layer, we solved the problem of breaking the IOB constraints. By learning to optimally combine the model predictions, EnsembleCRF also learns to avoid mistakes made by single sequence labeling models.

We experimented with creating $D_{en}$ with the augmented training dataset. However, we found that there is no positive correlation between the number of models in $C$  and the macro-averaged F1 score on the development dataset. Treating every possible combination of set $C$ as a hyperparameter to optimize will yield the optimal result. 

\subsubsection{Ensemble model accuracy}

In the MultiCONER competition \cite{multiconer-data, multiconer-report}, an ensemble model provided a boost in the F1 score. We followed a similar approach and used the same architecture. The single models, which were evaluated in the earlier section, are combined to form an ensemble. The results are tabulated in Table \ref{tab3}. We found all six models with 4 augmentations gave an F1 score of 0.9173 on the development set, but the best model evaluated on the test set had a different combination, which is represented, in bold format. The results varied at the third decimal digit position indicating that the combination is providing improvement but is not sufficient to bring in the change in the f1 score. However, the results are consistent with the ensemble model. One of the main concerns with the ensemble model is the number of single models to use to gain the maximum F1 score. Since it has to be evaluated in a trial-and-error approach.

\begin{table*}[h]
    \begin{center}
    \caption{The performance of ensemble with different text embeddings on the development set.}
    \label{tab3}
    \begin{tabular}{l|l|l} 
      \hline
      Model& Text embeddings & Development\\
      number & & F1 score\\
      \hline
      \hline
      1 & roberta-base + 4 augmentations & 0.9154\\
      (3 models) & roberta-base + glove + 4 augmentations & \\
      & xlm-roberta-base + 4 augmentations & \\
      \hline
      2 & roberta-base + 4 augmentations & 0.9164\\
      (3 models) & roberta-base + glove + 4 augmentations & \\
      & roberta-base + glove + nobilstm + 4 augmentations & \\
      \hline
      3 & roberta-base + 4 augmentations & 0.9161\\
      (4 models) & roberta-base + glove + 4 augmentations & \\
      & roberta-base + glove + nobilstm + 4 augmentations & \\
      & roberta-base + glove + nobilstmcrf + 4 augmentations & \\
      \hline
      4 & \textbf{roberta-base + 4 augmentations} & \textbf{0.9167}\\
      (5 models) & \textbf{roberta-base + glove + 4 augmentations} & \\
      & \textbf{roberta-base + glove + nobilstm + 4 augmentations} & \\
      & \textbf{roberta-base + glove + nobilstmcrf + 4 augmentations} & \\
      & \textbf{xlm-roberta-base + 4 augmentations} & \\
      \hline
      5 & roberta-base + 4 augmentations & 0.9173\\
      (6 models) & roberta-base + glove + 4 augmentations & \\
      & roberta-base + glove + nobilstm + 4 augmentations & \\
      & roberta-base + glove + nobilstmcrf + 4 augmentations & \\
      & xlm-roberta-base + 4 augmentations & \\
      & glove + 4 augmentations & \\
      \hline
    \end{tabular}
    \end{center}
\end{table*}

\subsection{Results of sub-task 1}

The test results are tabulated in Table \ref{tab4}. Our ensemble approach is placed at the top with an F1 score of 0.939. The results are tabulated after performing a reproducibility test by the competition organizers which helps us confirm that our programs are reproducible and consistent with the F1 score.

\begin{table*}[h]
    \begin{center}
    \caption{The performance of different submissions on the test set \cite{nl4optcomp}.}
    \label{tab4}
    \begin{tabular}{l|l|l|l} 
      \hline
      Rank & Team Name & Affiliation(s) & F1 score\\
      \hline
      \hline
      \textbf{1} & \textbf{Infrrd AI Lab} & \textbf{Infrrd} & \textbf{0.939}\\
      \hline
      2 & mcmc & OPD & 0.933\\
      \hline
      3 & PingAn-zhiniao & PingAn Technology & 0.932\\
      \hline
      4 & Long & BDAA-BASE & 0.931\\
      \hline
      5 & VTCC-NLP & Viettel & 0.929\\
      \hline
      6 & Sjang & POSTECH & 0.927\\
      \hline
      7 & DeepBlueAI & DeepBlueAI & 0.921\\
      \hline
      8 & TeamFid & Fidelity & 0.920\\
      \hline
      9 & KKKKKi & Netease & 0.917\\
      \hline
      10 & holajoa & Imperial College London & 0.910\\
      \hline
      11 & Dream & & 0.884\\
      \hline
      & Baseline (xlm-roberta-base) & Nl4Opt & 0.906\\
      \hline
    \end{tabular}
    \end{center}
\end{table*}

\section{Sub-task 2: text generation}

The generation of declarations was conceptualized as entity relationship mapping. The relationship between the ground truth entities was missing, and our refined objective was to map the relationship to capture the declaration that exists in the natural language. To do so, we tried to implement the entity relationship way of mapping the entities. We couldn’t progress further and moved to the next step, the text generation approach. We observed that the text generation process was much easier than entity relationship mapping in our experiments.

We use a text generator to generate the declarations. We have used text to text transfer transformer (T5) in our experiments \cite{t5}. The architecture has visible tokens for the input text, and the output text is observed based on the past predicted outputs. We fine-tuned the regularly available T5 model from the hugging face library \cite{huggingface}. In the first stage, we provided the raw input text as input to the transformer, and the expected output was as defined in sub-task 2 \cite{nl4optcomp}. Our output is a generation of declaration, and the results generated were on par with the baseline results shared by the competition organizers.

\begin{figure}[b]
\centering
\includegraphics[scale=1.5]{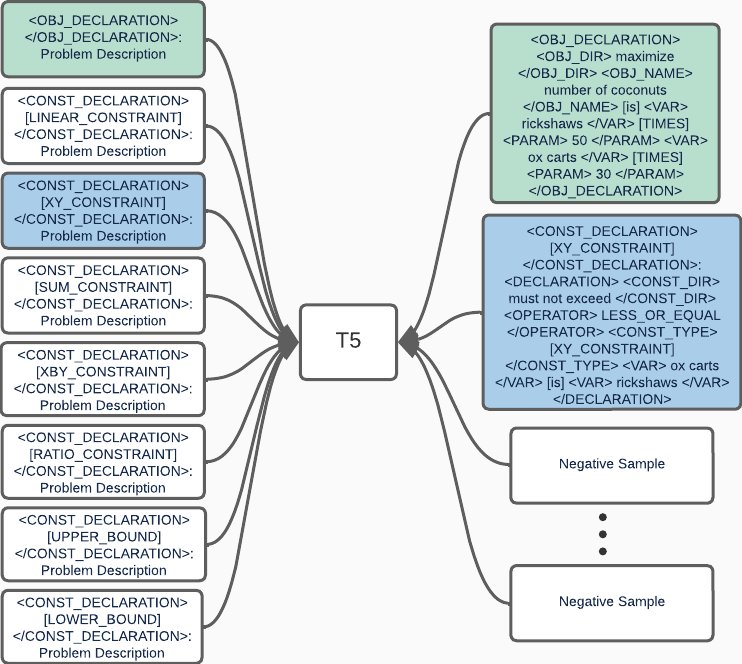}
\caption{The generation of text using T5 model for different types of prompts.}
\label{fig3}
\end{figure}

We observed that the score was par but didn’t have much-anticipated improvements. We performed an initial analysis and found the two points identified during our discussion. They were the number of samples available to train the model and the repeated occurrence of declarations in terms of constraints. To address these points, we decompose each sample into pieces for multi-task learning and, thus, implicitly increase the number of training samples in our dataset.

We believe the output was low due to the small number of samples. We split the single task into multiple tasks as shown in Figure \ref{fig3}. In this way, we invariably introduced a lot of examples in the training set. This approach helped the model to map the declaration with the probable entities in the process. First, we trained the model to predict only the objective of the problem, and it was then scaled to generate the constraint declarations. There were seven different types of constraints in the competition. They are sum constraints, upper bound constraints, lower bound constraints, linear constraints, ratio constraints, xby constraints, and xy constraints. Thus, we combine all the declarations by repeating the training seven more times.

Every linear programming problem has a definite objective statement, but the constraints declarations may be missing in a few. In our multi-task learning setting, the missing constraint statements in a problem act as negative samples in the training process. It helps the model in predicting the correct constraints without much effort. Tables \ref{tab5} and \ref{tab6} show the original mapping and the proposed mapping for sub-task 2 to add multiple samples to the training dataset.

\begin{table*}[h]
    \begin{center}
    \caption{The original mapping for text generation in sub-task 2 \cite{nl4optcomp}.}
    \label{tab5}
    \begin{tabular}{|l|} 
      \hline
      \textbf{Original mapping:}\\
      \hline
      \hline
\verb|<s>|\\
\verb|<DECLARATION>|\\
\verb|<OBJ_DIR> maximize </OBJ_DIR>|\\
\verb|<OBJ_NAME> number of coconuts </OBJ_NAME> [is]|\\
\verb|<VAR> rickshaws </VAR> [TIMES] <PARAM> 50 </PARAM>|\\
\verb|<VAR> ox carts </VAR> [TIMES] <PARAM> 30 </PARAM>|\\
\verb|</DECLARATION>|\\
\verb|<DECLARATION>|\\
\verb|<CONST_DIR> at most </CONST_DIR>|\\
\verb|<OPERATOR> LESS_OR_EQUAL </OPERATOR>|\\
\verb|<LIMIT> 200 </LIMIT>|\\
\verb|<CONST_TYPE> [LINEAR_CONSTRAINT] </CONST_TYPE> [is]|\\
\verb|<VAR> rickshaws </VAR> [TIMES] <PARAM> 10 </PARAM>|\\
\verb|<VAR> ox carts </VAR> [TIMES] <PARAM> 8 </PARAM>|\\
\verb|</DECLARATION>|\\
\verb|<DECLARATION>|\\
\verb|<CONST_DIR> must not exceed </CONST_DIR>|\\
\verb|<OPERATOR> LESS_OR_EQUAL </OPERATOR>|\\
\verb|<CONST_TYPE> [XY_CONSTRAINT] </CONST_TYPE>|\\
\verb|<VAR> ox carts </VAR> [is] <VAR> rickshaws </VAR>|\\
\verb|</DECLARATION>|\\
\verb|</s>|\\
\hline
    \end{tabular}
    \end{center}
\end{table*}

\begin{table*}[h]
    \begin{center}
    \caption{The exploratory data analysis of number of tokens for different training strategies.}
    \label{tab7}
    \begin{tabular}{l|l|l} 
      \hline
      Type of training & Prefix + input length (max) & Output mapping length (max)\\
      \hline
      \hline
      Original: prefix=generate & Train: 203 & Train: 716\\
      linear program mapping & Dev: 214 & Dev: 601\\
      \hline
      augmented input & Train: 545 & Train: 716\\
      & Dev: 529 & Dev: 601\\
      \hline
      multi-task & Train: 248 & Train: 564\\
      & Dev: 258 & Dev: 438\\
      \hline
      augmented input + multi-task & Train: 591 & Train: 564\\
      & Dev: 570 & Dev: 438\\
\hline
    \end{tabular}
    \end{center}
\end{table*}

\begin{table*}[ht]
    \begin{center}
    \caption{The proposed multi-task mapping for text generation.}
    \label{tab6}
    \begin{tabular}{|l|} 
      \hline
      \textbf{Multi-task mapping:}\\
      \hline
      \hline
\verb|prompt <OBJ_DECLARATION> </OBJ_DECLARATION>:|\\
\verb|<OBJ_DECLARATION>|\\
\verb|<OBJ_DIR> maximize </OBJ_DIR>|\\
\verb|<OBJ_NAME> number of coconuts </OBJ_NAME> [is]|\\
\verb|<VAR> rickshaws </VAR> [TIMES] <PARAM> 50 </PARAM>|\\
\verb|<VAR> ox carts </VAR> [TIMES] <PARAM> 30 </PARAM>|\\
\verb|</OBJ_DECLARATION>|\\
\\
\verb|prompt <CONST_DECLARATION> [LINEAR_CONSTRAINT] </CONST_DECLARATION>:|\\
\verb|<CONST_DECLARATION>|\\
\verb|<CONST_DIR> at most </CONST_DIR>|\\
\verb|<OPERATOR> LESS_OR_EQUAL </OPERATOR>|\\
\verb|<LIMIT> 200 </LIMIT>|\\
\verb|<CONST_TYPE> [LINEAR_CONSTRAINT] </CONST_TYPE> [is]|\\
\verb|<VAR> rickshaws </VAR> [TIMES] <PARAM> 10 </PARAM>|\\
\verb|<VAR> ox carts </VAR> [TIMES] <PARAM> 8 </PARAM>|\\
\verb|</CONST_DECLARATION>|\\
\\
\verb|prompt <CONST_DECLARATION> [XY_CONSTRAINT] </CONST_DECLARATION>:|\\
\verb|<DECLARATION>|\\
\verb|<CONST_DIR> must not exceed </CONST_DIR>|\\
\verb|<OPERATOR> LESS_OR_EQUAL </OPERATOR>|\\
\verb|<CONST_TYPE> [XY_CONSTRAINT] </CONST_TYPE>|\\
\verb|<VAR> ox carts </VAR> [is] <VAR> rickshaws </VAR>|\\
\verb|</DECLARATION>|\\
\\
\textbf{negative samples}\\
\verb|prompt <CONST_DECLARATION> [SUM_CONSTRAINT] </CONST_DECLARATION>:|\\
\verb|prompt <CONST_DECLARATION> [XY_CONSTRAINT] </CONST_DECLARATION>:|\\
\verb|prompt <CONST_DECLARATION> [RATIO_CONSTRAINT] </CONST_DECLARATION>:|\\
\verb|prompt <CONST_DECLARATION> [UPPER_BOUND] </CONST_DECLARATION>:|\\
\verb|prompt <CONST_DECLARATION> [LOWER_BOUND] </CONST_DECLARATION>:|\\
\hline
    \end{tabular}
    \end{center}
\end{table*}

We fine-tune the model through prompting. We prefix the input text to generate the appropriate output depending on the prefix task. We performed exploratory data analysis to compute the maximum number of tokens after prefix at the input and the output. The numbers are tabulated in Table \ref{tab7} and exceeded the limit of 512 tokens in the original training set. After splitting the main task into multiple tasks and increasing the data, we observed that the results started moving up and away from the baseline results. The results were not as expected after this step, and another round of analysis was carried out to identify the reasons for not picking up the correct declarations from the statements.

\begin{figure}[b]
\centering
\includegraphics[scale=1.5]{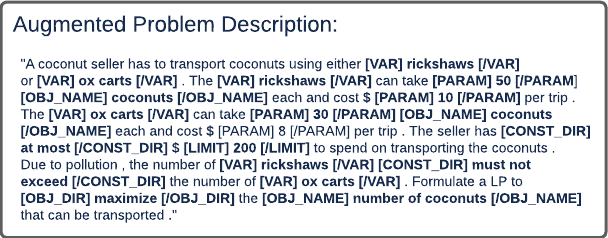}
\caption{An example word problem with the wrapped entities used in the training.}
\label{fig4}
\end{figure}

\begin{table*}[t]
    \begin{center}
    \caption{The declaration-level accuracy on the development set for different types of training with the hyperparameters.}
    \label{tab8}
    \begin{tabular}{l|l|l} 
      \hline
      Type of training & Hyperparameters & Declaration-level accuracy\\
      \hline
      \hline
      Original: prefix=generate & max\textunderscore seq\textunderscore length: 600 & 62.8\%\\
      linear program mapping & max\textunderscore output\textunderscore length: 750 & \\
      & training\textunderscore epochs: 20 & \\
      \hline
      augmented input & max\textunderscore seq\textunderscore length: 600 & 75.9\%\\
      & max\textunderscore output\textunderscore length: 750 & \\
      & training\textunderscore epochs: 24 & \\
      \hline
      multi-task & max\textunderscore seq\textunderscore length: 600 & 65.6\%\\
      & max\textunderscore output\textunderscore length: 750 & \\
      & training\textunderscore epochs: 17 & \\
      \hline
      augmented input + multi-task & max\textunderscore seq\textunderscore length: 600 & 83.5\%\\
      & max\textunderscore output\textunderscore length: 750 & \\
      & training\textunderscore epochs: 50 & \\
\hline
    \end{tabular}
    \end{center}
\end{table*}

We thought of teaching the model by explicitly mentioning the actual locations of the entities in the input text. Thus, we prepared a wrapper for all the entities with a notation shown in Figure \ref{fig4}. The entities present in the input text get wrapped before training the model. The number of tokens doubled with the addition of the wrapper tokens. Table \ref{tab7} shows the increase in the number of tokens processed by the model. We observed the maximum number of tokens in the training set before and after the addition of the wrapper are 240 and 590, respectively. With the addition of new tokens, the sequence length crossed the actual limit of 512, as suggested in the paper \cite{t5}. We have to fix the size of the transformer to 600 and perform training on the model. Since the size is not a power of 2, the performance deteriorates to some extent. Table \ref{tab8} shows the ablation study for different training strategies.

\subsection{Results of sub-task 2}

The results of sub-task 2 on the test set is tabulated in Table \ref{tab9}. We stand in the fifth position after evaluating the program independently by the competition organizers and thus results are updated by reproducing the model by the organizers. There is a slight drop in the accuracy due to the retrieval of multiple instances for the same prompt question.

\begin{table*}[h]
    \begin{center}
    \caption{The performance of different submissions on the test set \cite{nl4optcomp}.}
    \label{tab9}
    \begin{tabular}{l|l|l|l} 
      \hline
      Rank & Team Name & Affiliation(s) & F1 score\\
      \hline
      \hline
      1 & UIUC-NLP & UIUC & 0.899\\
      \hline
      2 & Sjang & POSTECH & 0.878\\
      \hline
      3 & Long & BDAA-BASE & 0.867\\
      \hline
      4 & PingAn-zhiniao & PingAn Technology & 0.866\\
      \hline
      \textbf{5} & \textbf{Infrrd AI Lab} & \textbf{Infrrd} & \textbf{0.780}\\
      \hline
      6 & KKKKKi & Netease & 0.634\\
      \hline
      & Baseline (BART) & NL4Opt & 0.608\\
      \hline
    \end{tabular}
    \end{center}
\end{table*}

\subsection{Discussion}

The additional wrapper tokens not only increase the length of the input sequence and redundant for many repeated items. We have to replace each wrapper token with a unique token to limit the size of tokens doesn’t cross the prescribed limit of 512. We have performed data augmentation in sub-task 1. We may need to use this approach to increase the number of samples in the training set. The objective and constraints declarations consist of a lot of numerical values. It is difficult to identify whether to perform the replacement of a numerical value or word reflecting the entity in a problem. The retrieval of the model through prompting has one serious concern to be addressed. When the entity appears at multiple locations, the model may pick one of them and predict it as an output. If we have different values at several locations for the same entity, the model picks one and doesn’t retrieve all of them. The values present in the input text differ for the same entity. It has to be addressed in the future to make the model adapt to picking up all of them belonging to the same entity in the training procedure.

\section{Conclusion}

We proposed an ensemble of models to predict the labels and generate the texts. The approaches are consistent with better results compared to a single model. We would like to explore the decoders in the ensemble since the text generation had the decoder part. We would like to use the large models in place of the base models to know whether the models are better at predicting the labels. The OBJ NAME entity is difficult to capture in the present set of models which has to be addressed in a different way. We need to overcome the drawbacks of prompting like two instances in the same question to improve the score.

\bibliographystyle{plainnat}
\bibliography{custom}

\end{document}